\documentclass[conf]{new-aiaa}
\usepackage[utf8]{inputenc}
\usepackage{graphicx}
\usepackage{amsmath}
\usepackage[version=4]{mhchem}
\usepackage{siunitx}
\usepackage{longtable,tabularx}
\usepackage{algorithm}
\usepackage{algpseudocode}
\usepackage{booktabs}

\usepackage{tikz}
\usetikzlibrary{shapes.geometric, arrows.meta, positioning, calc, fit, backgrounds, shadows}

\usepackage{microtype}
\usetikzlibrary{arrows.meta, positioning}%
\usepackage{subcaption}  %
\usepackage{xcolor}
\title{A Future Capabilities Agent for Tactical Air Traffic Control}

\author{
Paul Kent, George De Ath, Martin Layton, Allen Hart\footnote{Research Fellow (Kent, Layton, Hart), Lecturer (De Ath), and Professor (Everson), University of Exeter.}
}
\affil{University of Exeter, Exeter, UK}

\author{Richard Everson}

\affil{The Alan Turing Institute, London, UK \& University of Exeter, UK}

\author{
Ben Carvell\footnote{Lead Researcher, NATS.}
}
\affil{NATS, Whiteley, UK}

\begin{document}

\maketitle

\begin{abstract}

Escalating air traffic demand is driving the adoption of automation to support air traffic controllers, but existing approaches face a trade-off between safety assurance and interpretability. Optimisation-based methods such as reinforcement learning offer strong performance but are difficult to verify and explain, while rules-based systems are transparent yet rarely check safety under uncertainty. This paper outlines Agent Mallard, a forward-planning, rules-based agent for tactical control in systemised airspace that embeds a stochastic digital twin directly into its conflict-resolution loop. Mallard operates on predefined GPS-guided routes, reducing continuous 4D vectoring to discrete choices over lanes and levels, and constructs hierarchical plans from an expert-informed library of deconfliction strategies. A depth-limited backtracking search uses causal attribution, topological plan splicing, and monotonic axis constraints to seek a complete safe plan for all aircraft, validating each candidate manoeuvre against uncertain execution scenarios (e.g., wind variation, pilot response, communication loss) before commitment. 

Preliminary walkthroughs with UK controllers and initial tests in the BluebirdDT airspace digital twin indicate that Mallard's behaviour aligns with expert reasoning and resolves conflicts in simplified scenarios. The architecture is intended to combine model-based safety assessment, interpretable decision logic, and tractable computational performance in future structured en-route environments.
\end{abstract}
\section{Introduction}
The imperative to manage escalating air traffic worldwide~\citep{FAA2024AerospaceForecastTables, EUROCONTROL2018ChallengesGrowth} is driving Air Navigation Service Providers to increasingly adopt automated tools and explore the potential of artificial intelligence (AI) to enhance safety, efficiency, and scalability~\citep{NATS2014iFACTS, FAA2014URET, EUROCONTROL2017MTCD, NASA2019ATD2, SESAR2022AISA, Turing2025Bluebird}. These efforts aim to support Air Traffic Control Officers (ATCOs) as they navigate increasingly complex operational environments. 

Recent research has explored advanced AI techniques, particularly multi-agent reinforcement learning, to address Air Traffic Control (ATC)  complexity~\citep{Ghosh2020DeepEnsembleMARL, Brittain2019AutonomousATC, Nilsson2025SelfPrioritizingMARL}. However, deep learning-based approaches are often criticised for their opaque, black-box nature, hindering human trust and practical deployment in safety-critical environments~\citep{SESAR2022AISA, EASA2020AIRoadmap}. Additionally, these models frequently rely on idealised assumptions (e.g., perfect information, unconstrained manoeuvrability), creating a significant gap between theoretical performance and operational reality~\citep{IFATCA2022AIChallengesATM, PerezCerrolaza2023AISafetyCritical}. Addressing this gap requires principled simplifications of operational complexity, aligning algorithmic development with realistic operational evolutions.

To bridge this complexity–transparency gap, this work outlines Agent Mallard, a forward-planning, rules-based AI agent explicitly designed around the concept of systemised airspace~\citep{NATS2025FutureAirspace}. Mallard is intended to exploit the inherent structure and predictability provided by predefined GPS-guided routes to simplify tactical decision-making and conflict resolution. This structured approach is designed to support transparency and verifiability, which are critical for human trust, while providing a pathway to advanced decision-support capabilities within future operational airspace configurations.

\subsection{Contributions}

This work makes the following contributions to autonomous air traffic control research:

\begin{enumerate}
    \item \textbf{Simulation-Integrated Architecture:} We propose an agent architecture that evaluates every candidate control action through forward simulation under uncertainty before execution. This design aims to provide model-based safety assessment, including stochastic factors such as speed uncertainty, and interpretability, properties that are often traded off in existing approaches.
    
    \item \textbf{Coordinated Conflict Resolution Algorithm:} We introduce an airspace-wide resolution algorithm that seeks conflict-free trajectories for all aircraft in a sector simultaneously rather than resolving pairwise conflicts in isolation. The algorithm employs a depth-limited, backtracking search with specialised pruning mechanisms to manage cascading interactions efficiently.
    
    \item \textbf{Structured Airspace Design Principles:} We show how predefined GPS-guided route structures can reduce the conflict-resolution problem from continuous trajectory optimisation to discrete choice selection, supporting tractable real-time performance with explicit safety assessment via digital-twin simulation.
\end{enumerate}

Preliminary validation results from expert controller reviews and initial digital 
twin testing are presented in Section~\ref{sec:current_status}, with complete evaluation 
through an agent validation framework identified as ongoing work.

The remainder of the paper is as follows: Section~\ref{sec:background} reviews air traffic control operations and related approaches to autonomous conflict resolution. Section~\ref{sec:mallard} presents the Agent Mallard architecture, including hierarchical planning, conflict resolution algorithms, and interpretability. Section~\ref{sec:validation} describes the proposed validation methodology, while Section~\ref{sec:current_status} reports preliminary results from expert reviews and initial testing. Section~\ref{sec:assumptions_limitations} discusses design assumptions and known limitations, and Section~\ref{sec:future_work} identifies priorities for future development.

\section{Background and related work}\label{sec:background}
\subsection{Air Traffic Control and the drive for automation}

En-route air traffic control manages aircraft during climb, cruise, and descent 
across multiple sectors. Controllers maintain separation, sequence traffic, and 
coordinate handovers for the sector of airspace they are responsible for.  This imposes a high cognitive load \citep{Suarez2024_ATCWorkloadReview, histon2008mitigating, Djokic2010_ATCcomplexityWorkload}. As traffic densities 
grow, automated monitoring can reinforce controller capacity by maintaining 
continuous surveillance, reserving human expertise for strategic coordination.

\subsection{Systemised Airspace: A Foundation for Future ATC}
A key ingredient to enabling such automation is the modernisation of the airspace structure itself. Systemised airspace marks a major shift in air traffic management, replacing tactical vectoring with predefined, GPS-guided routes known as ``motorways in the sky''~\citep{NATS2025FutureAirspace}. Enabled by Performance-Based Navigation (PBN) standards such as RNAV1~\citep{ICAO2014RNAVIntro}, this structure allows aircraft to follow end-to-end trajectories with minimal controller intervention~\citep{FAA2022NextGenAnnualReport}.

This global shift is led by major modernisation efforts. In Europe, SESAR promotes systemised, trajectory-based operations, including Point Merge and widespread PBN route deployment~\citep{EUROCONTROL2025PointMerge, SESAR2023BAWP}. In the U.S., the FAA's NextGen programme has rolled out PBN-based standard instrument departures (SIDs) and arrivals (STARs) through its Metroplex initiative~\citep{FAA2022NextGenAnnualReport}.

Using AI agents for tactical ATC is challenging in part due to the large number of actions that an aircraft can take at any moment.  For example, headings are usually discretised to $5^\circ$ increments, so an aircraft might be vectored in any one of $360/5 = 70$ directions and might climb or descend to any one of the 20 flight levels between 20,000 ft and 40,000 ft. When two aircraft are involved the dimension of the joint action space may exceed 100.  Structured airspace offers a unique opportunity for AI. By reducing complexity through predefined geometries, it enables simpler, more transparent algorithms that are better aligned with human oversight and operational practice, and more suitable for human-machine collaborative, safety-critical environments.

PBN greatly improves horizontal navigation reliability, continually correcting for wind and drift to keep aircraft close to their intended ground track. This reduces lateral uncertainty to a bounded cross-track error and shifts most variability into along-track timing. Systemised airspace is designed to exploit this property: by placing routes with fixed geometric spacing, lateral separation can be maintained through geometry rather than continuous tactical vectoring. As a result, the conflict-resolution problem can be simplified from full 3D vectoring to managing vertical transitions and longitudinal spacing in suitably structured sectors.

To realise these geometric guarantees in practice, systemised sectors must provide a lane structure whose spacing is preserved along the entire route. Figure~\ref{fig:lanes} illustrates how this is implemented in the BluebirdDT digital twin, where fix-to-fix routes are accompanied by adjacent deconfliction lanes. The lane structure is constructed geometrically: parallel lines are placed 3.5~NM on either side of the centreline connecting successive fixes along the route. At route turns, these offset lines intersect to form transition nodes. This construction preserves the 7~NM spacing (3.5~NM either side) throughout the route, including through turns, ensuring that separation criteria are maintained regardless of an aircraft’s along-track position.

\subsection{Approaches to Autonomous Conflict Resolution}

\subsubsection{Optimisation and Geometric Approaches}
Early approaches to autonomous conflict resolution framed the problem as a global optimisation, utilising mixed-integer programming or evolutionary algorithms (e.g., genetic algorithms) to search for conflict-free 4D trajectories \citep{Delahaye_Genetic, Durand_Optimal}. While theoretically capable of finding optimal solutions, these methods are computationally intensive, often struggling to meet real-time constraints in high-density sectors. Furthermore, although hard constraints like sector boundaries can be captured mathematically, encoding the softer stylistic preferences of ATC within a cost function remains extremely challenging. Controllers expect patterns such as standard flows, tidy headings, and manoeuvres that look natural. Optimisation methods frequently produce trajectories that satisfy the mathematical constraints but look erratic or robotic to human operators, reducing trust and making it harder to express the intended manoeuvre using standard voice phraseology.

\subsubsection{Reinforcement Learning}
Some recent efforts have focused on Reinforcement Learning (RL), where agents learn deconfliction policies through trial and error in simulated environments \citep{Brittain2019AutonomousATC, isufaj2022toward}. While RL and multi-agent RL (MARL) have demonstrated impressive scalability, it introduces significant challenges for safety-critical deployment. The learned policies are typically opaque ``black boxes,'' hindering verification and building of trust~\citep{bluebird_agent_transparency}. Additionally, continuous-action RL agents often exhibit high-frequency control jitter (issuing micro-adjustments to heading or speed) which creates excessive pilot workload and violates the discrete nature of ATC voice communications. One recent technique, the Online Action-Stacking method of \citet{bluebird_rl},  mitigates this issue by aggregating micro-actions into single instructions. Yet, the training of MARL agents frequently necessitates non-operational simplifications to the environment to facilitate convergence. Consequently, a significant gap remains between the theoretical performance of these models and the safety assurance required for operational deployment.

\subsubsection{Rule-Based and Heuristic Agents}
Rule-based and heuristic agents approach air traffic control by following predefined logic, often in the form of expert-designed \textit{if-then} rules. Early systems in this category sought to directly encode controller knowledge, while later developments introduced more sophisticated cognitive models. A notable example is NASA's CATS-based agent \citep{callantine2002cats}, which structured controller behaviour as a hierarchy of intent-driven tasks (such as monitoring, conflict detection, and manoeuvre execution) designed to emulate human decision-making.

Subsequent NASA research extended these agents into Monte Carlo simulation frameworks like TRAC, enabling large-scale testing of procedural robustness under varied traffic and weather conditions \citep{callantine2009trac}. However, in these systems, simulation was used purely as an \textit{offline} analysis tool—not as a core part of the agent's real-time decision logic.

More recently, systems like the Bluebird expert rules-based agent~\citep{bluebird_agent_transparency} which apply a fixed set of reactive heuristics with short-horizon simulation, have shown that simple reactive rules can yield promising results with low computational cost. These agents are easy to interpret and closely mirror controller workflows. However, they generally depend on \textit{procedural validity}—the assumption that applying a standard manoeuvre in a known geometry will produce a safe outcome. Without extended-horizon verification integrated into the planning loop, this assumption may break down in complex scenarios: resolving a local conflict may inadvertently create a new, more severe interaction downstream \citep{prevot2002exploring}.

\subsubsection{The Gap: Simulation-Integrated Decision-Making}
A critical gap persists in agent design for ATC: systems must verify safety under uncertainty while remaining interpretable to human operators. Current approaches tend to make trade-offs—optimisation-based agents (e.g., RL or evolutionary planners) offer robustness but lack transparency; rule-based systems are interpretable but rarely verify outcomes through simulation; and architectures like CATS/TRAC treat simulation as an offline tool, disconnected from real-time decision-making.
Agent Mallard is intended to help close this gap by embedding stochastic forward simulation directly into its manoeuvre-selection loop. Like CATS, it issues high-level, intent-based actions structured in a hierarchical decision process, preserving compatibility with controller workflows. But unlike earlier agents, Mallard is designed to use its digital twin to simulate each candidate manoeuvre before issuing it — checking for secondary conflicts and constraint violations across a range of uncertain execution scenarios, including wind variation and pilot behaviour.

Importantly, Mallard aims to do this without incurring full 4D optimisation overhead by leveraging systemised airspace. Predefined lateral routes (e.g., PBN lanes) reduce the solution space, allowing the agent to focus on vertical and temporal adjustments. The overall design is intended to combine three properties that are often traded off in prior work: interpretability, robustness assessed via stochastic simulation, and the potential for real-time performance in structured en-route environments.

\section{Agent Mallard: Proposed Approach} 
\label{sec:mallard}
\begin{figure}
\centering
\includegraphics[width=\linewidth]{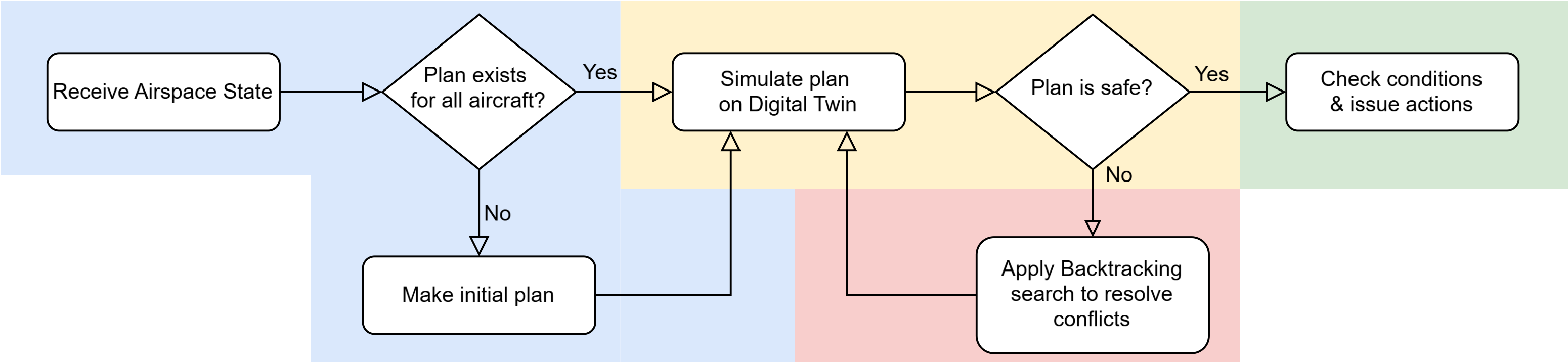}
\caption{Agent Mallard's operational flowchart. Blue: initialisation (airspace state received, plans generated if needed). Yellow: evaluation (plan simulated on the digital twin). Red: deconfliction (if unsafe, Backtracking search looks for a safe alternative; see Sec.~\ref{sec:backtracking}). Green: execution (if safe, conditions checked and actions issued). If conditions are never met the aircraft would miss its coordination, be deemed unsafe, and planned again.
\label{fig:flow}}
\end{figure}
Agent Mallard is a forward-planning agent designed to provide tactical air traffic control in systemised airspace. The agent is designed to continuously monitor aircraft, predict conflicts through forward simulation, and generate sets of clearances that resolve all predicted safety violations simultaneously.

Mallard's design rests on a straightforward principle: \textit{simulate before you commit}. Every candidate solution is tested against a high-fidelity digital twin of the airspace before acceptance. The simulation is capable of accounting for real-world uncertainties: wind, pilot response times and, critically, communication loss. Only solutions that maintain safe separation against all these issues are accepted as valid.

Mallard issues commands in the same manner as human controllers: standard clearances for climbs, descents, and speed adjustments. In systemised airspace, lateral control is simplified: rather than issuing arbitrary heading vectors, the agent directs aircraft to follow specific predefined lanes, the GPS-guided routes that form the airspace structure. This lane-based approach reduces the lateral control problem from continuous vectoring to discrete lane selection, enabling safety verification through simulation and maintaining geometric separation by design. It also reduces the cognitive load on controllers, since aircraft remain on predictable paths that require less monitoring while still guaranteeing safe separation.

When conflicts are found, through simulation, Mallard uses its deconfliction logic to identify the type of conflict and then propose a strategy. Mallard does this from a list of deconfliction strategies, attempting to identify the cause, and suggest a cure. Because every decision traces back to a recognisable tactical pattern applied to a specific conflict geometry, the agent's reasoning remains transparent: human operators can understand not just \textit{what} clearances are proposed, but \textit{why} that particular approach was selected.

\subsection{How Mallard Operates: A Complete Cycle}

Figure~\ref{fig:flow} illustrates Mallard's operational cycle, which repeats every few seconds throughout operations: \\

\textbf{Baseline Planning (Blue):} The agent receives the current airspace state—positions, velocities, and flight plans for all aircraft in the sector plus expected arrivals within 15 minutes. For each aircraft, Mallard constructs a baseline plan representing the most efficient path: climb to preferred altitude, follow the assigned GPS-guided route, descend at the optimal point to meet coordinated exit conditions. These plans assume no conflicts and maximise operational efficiency.

\textbf{Safety Verification (Yellow):} Mallard tests these plans by forward-simulating the next hour of operations. The digital twin rolls forward all aircraft trajectories under multiple scenarios: nominal conditions, realistic wind variations, delayed pilot responses, and a 15-minute loss-of-communication scenario where aircraft continue their last clearance with no further instructions. This verification reveals both immediate conflicts and downstream interactions that might emerge minutes into the future.

Mallard resolves conflicts sequentially, treating each loss of separation as a two-aircraft problem. For each conflict, the agent assigns complementary clearances to both aircraft (e.g., both offset left to parallel lanes, or one climbs while the other descends). Each candidate strategy is injected into the corresponding flight plans, after which the entire airspace is re-simulated to assess downstream effects and verify that no new conflicts are introduced: perhaps the newly offset aircraft now conflicts with a third aircraft on an adjacent route, or the timing change alters a follow-on interaction. If new conflicts appear, they are addressed in turn through the same process: identify the responsible plan segments, evaluate strategies, re-simulate. The search continues recursively, potentially trying strategy combinations at increasing depths, until a complete solution is found, a coordinated airspace plan in which all aircraft maintain safe separation simultaneously. Throughout this process, only the specific plan segments causing conflicts are modified; the rest of each aircraft's efficient baseline trajectory remains intact.

\textbf{Clearance Issuance (Green):} Once a verified safe solution exists, Mallard monitors the airspace and issues clearances at the appropriate moments. Clearances are bound to geometric conditions rather than fixed times, e.g., ``turn left when lateral separation exceeds 5 nautical miles'', ensuring safety even when timing varies. This state-based approach means clearance logic adapts automatically to actual trajectories rather than predicted schedules.

\textbf{Continuous Adaptation:} This cycle: \textit{plan, verify, resolve, issue}, repeats every few seconds. As aircraft move, new flights enter, and conditions change, Mallard re-evaluates through forward simulation and adapts accordingly.

\begin{figure}[t]
    \centering
    \includegraphics[width=\linewidth]{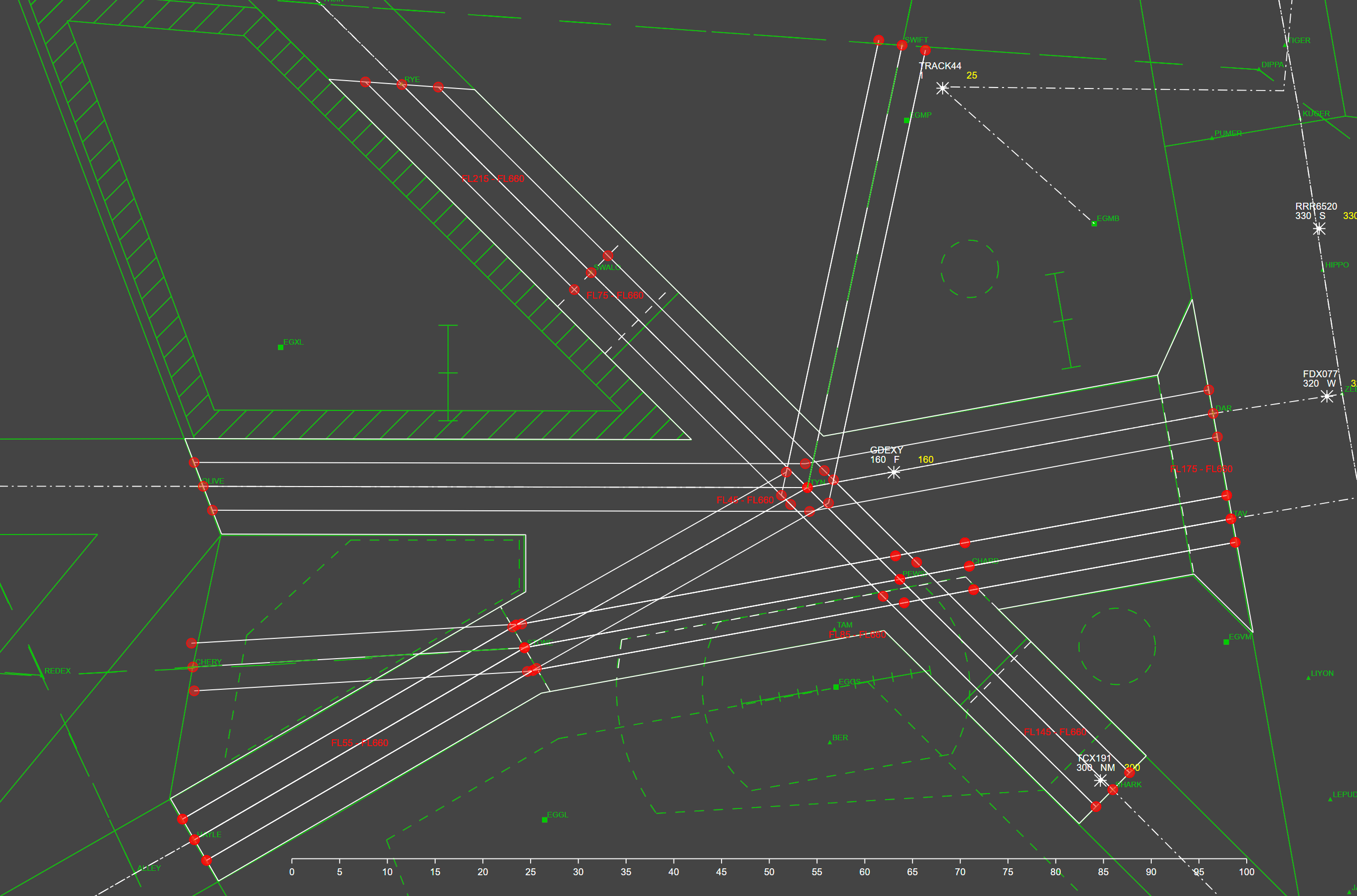}
    \caption{A set of deconflicted lanes for a training sector. The Right and Left lanes for each route are geometrically deconflicted from each other.}
    \label{fig:lanes}
\end{figure}

\subsection{Key components}
The agent Mallard methodology integrates five key components:
\begin{itemize}
\item \textbf{Systemised Airspace Structure:} Mallard exploits predefined GPS-guided route structures that form ``motorways in the sky'' (see Figure~\ref{fig:lanes}.) By constraining aircraft to follow discrete lateral routes with known geometric spacing, the lane-based structure reduces the conflict-resolution problem from continuous 4D trajectory optimisation to discrete choice selection over lanes, levels and relative speed. 
    \item \textbf{Trajectory Simulation with Uncertainty:} A high-fidelity digital twin of the airspace simulates the evolution of aircraft trajectories under proposed control actions, enabling forward-looking assessment of safety and operational acceptability in the presence of environmental uncertainty.
    \item \textbf{State-Based Conditional Policies:} Control interventions are encoded as dynamic policies (detailed in Section~\ref{sec:splicing}) triggered by local airspace geometry (e.g., separation thresholds or encounter types). This state-based formulation decouples safety from exact timing, ensuring that intent remains valid even when stochastic factors, such as wind, introduce along-track variability. 
    \item \textbf{Expert-Informed Decision-Making:} Mallard employs a decision logic grounded in expert ATCO reasoning (Table~\ref{tab:strategy_library}). By constraining the solution space to a ranked set of standard deconfliction strategies, the iterative backtracking search produces plans that are inherently interpretable, allowing human operators to verify not just the outcome, but the strategic rationale behind every intervention.
    \item \textbf{Causal Traceability:} The hierarchical structure creates an explicit audit trail for every control intervention. Because every Action is inextricably linked to a specific Trigger Condition and a parent Manoeuvre, the system preserves the logical provenance of its decisions. This allows human operators to scrutinise not only \textit{what} command was issued, but precisely \textit{why} and \textit{when} the logic determined it was necessary, rendering the decision-making process transparent and verifiable (Section~\ref{sec:causal}, Figure~\ref{fig:causal_attribution}). 

\end{itemize}

The following sections will elaborate on the architecture and specifics of implementation of these components and discuss how such an agent's performance could be assessed.

\subsection{Real-time monitoring and trajectory prediction}\label{sec:digital_twin_description}

Agent Mallard incorporates the BluebirdDT Digital Twin platform, part of the UK's Project Bluebird~\citep{bluebird_dt, NATS2025ProjectBluebird}. BluebirdDT provides real-world operational data for grounded validation~\citep{bluebird_assurance} and incorporates elements of ATCO training~\citep{bluebird_validation}, allowing agent performance to be assessed using metrics derived from the legal requirements governing human ATCO assessment~\citep{bluebird_validation}. This integration is intended to help keep Mallard aligned not only with technical standards, but also with operational expectations in structured environments as the design matures.

During testing, Mallard forward-simulates the airspace, evaluating both nominal trajectories and stochastic counterfactuals. It includes all aircraft in the sector and expected entries within 15 minutes, enabling proactive detection of future interactions. A one-hour maximum horizon supports traceability, while each counterfactual is also rolled out for 15 minutes assuming no further instructions, providing a conservative loss-of-communication safety buffer if communications are lost.

These horizons are chosen to match typical sector look-ahead times and loss-of-communication procedures in en-route operations. Uncertainty in execution is represented by perturbing nominal trajectories with sampled variations in key factors such as aircraft performance and environmental conditions, as supported by the probabilistic models in BluebirdDT~\citep{bluebird_dt, bluebird_prob_tp, bluebird_tp_descents}. Each rollout thus corresponds to a plausible realisation of the same control plan under different execution conditions.

The following sections detail each component of this architecture.

\subsection{Hierarchical Plan Representation}\label{sec:hierarchy}
Mallard represents each aircraft’s intent as a structured hierarchy rather than a fixed sequence of coordinates. A plan encodes not only \textit{where} an aircraft should be, but \textit{what} actions it should take and \textit{when} those actions should trigger in response to the evolving airspace state.

This representation serves two key purposes. First, it enables reasoning at multiple levels of abstraction: the agent can adjust a high-level manoeuvre without reconstructing its constituent clearances, or modify a single planned action without disturbing the remainder of the flight plan. Second, it ensures that plans remain valid under execution uncertainty without requiring constant replanning.

Instead of specifying actions at fixed times, e.g. ``turn heading 270 at 14:23'', Mallard binds actions to geometric or state-based conditions, such as ``turn heading 270 when lateral separation exceeds 5 NM.'' Conditions determine \textit{when} an action should occur based on the actual airspace state rather than a predicted schedule. This has a crucial property: if wind, performance variation, or pilot delay alters the aircraft’s timing, the plan remains logically correct. The action simply triggers when its condition becomes true, rather than at a predetermined time that may no longer be appropriate. The trajectory unfolds differently in time, but the underlying intent remains intact. During Mallard’s continuous re-simulation of the sector, the same plan is re-validated under new conditions rather than reconstructed.

The hierarchy decomposes an aircraft’s complete plan into progressively simpler components (Table~\ref{tab:control_hierarchy}). At the sector level, an airspace plan coordinates all aircraft. Each aircraft has a flight plan: a sequence of manoeuvres spanning entry to exit. Manoeuvres represent multi-phase tactical solutions (e.g., ``offset left, maintain separation, resume route'') and decompose into Planned Actions—closed-loop trigger–action–completion units. At the foundation lie atomic Conditions and Actions, the Boolean predicates and control commands that form the building blocks of all decision logic. The following sections introduce these primitives and show how they combine to form interpretable, robust plans.

\begin{table}[tb]
    \centering
    \caption{Hierarchy of Control Structures. The architecture decomposes high-level intent into atomic primitives. Note that a \textit{Planned Action} is a container that binds a specific logic gate (Condition) to a specific control output (Action).}
    \label{tab:control_hierarchy}
    \renewcommand{\arraystretch}{1.4}
    \begin{tabular}{@{}p{0.12\linewidth}p{0.22\linewidth}p{0.30\linewidth}p{0.28\linewidth}@{}}
        \toprule
        \textbf{Layer} & \textbf{Control Object} & \textbf{Function} & \textbf{Example}\\ 
        \midrule

        \textbf{Atomic} & 
        \textit{Condition} \newline \& \textit{Action} & 
        The indivisible logic gates and control verbs that constitute the plan. & 
        $C =$ \texttt{LatSep > 5NM} \newline
        $A = $ \texttt{ResumeNav(Fix)} \\
        \midrule

        \textbf{Tactical} & 
        \textit{Planned Action} & 
        A container binding a trigger condition, an action and a completion condition together. & 
        $\mathcal{P} = \langle C_{\text{trigger}}, A, C_{\text{completion}} \rangle$ \\
        \midrule

        \textbf{Strategic} & 
        \textit{Manoeuvre } & 
        A multi-step solution selected from the expert library. A sequence of planned actions, often the completion condition of one planned action is the trigger for another & 
        \textbf{Strategy B}: \newline  
        $\left( \mathcal{P}_n \to \mathcal{P}_{n+1}  \right)  $\\
        \midrule

        \textbf{Global} & 
        \textit{Flight Plan} & 
        The per aircraft end-to-end plan, composed of a linked sequence of Manoeuvres. & 
        $FP = \left( \mathcal{P}_1 \to \mathcal{P}_{2}  ... \to \mathcal{P}_{N}\right)  $ \\
        \midrule
        \textbf{System-Wide} & 
        \textit{Airspace Plan} & 
        The coordinated set of conflict-free flight plans for all aircraft in the sector simultaneously. & 
        $S = \{FP_1, FP_2, ..., FP_M\}$ \\
        \bottomrule
    \end{tabular}
\end{table}

\subsubsection*{Fundamental Primitives: Conditions and Actions}
Mallard's control logic is built from two primitives: \textit{Conditions} and \textit{Actions}.  Actions are executed if and only if the associated condition is met. Conditions $(C)$ comprise Boolean predicates describing airspace states including temporal checks and relative geometric triggers such as \texttt{AircraftPassedLaterally}. This geometric formulation decouples safety from timing uncertainty, ensuring that intent remains valid even when stochastic factors introduce along-track variability. Actions $A$ are discrete ATC instructions expressed in standard phraseology, such as 
\textit{Climb FL330} or \textit{Fly Heading 270}.

\subsubsection*{Independent Control Axes}
In Mallard’s control model, the lateral and vertical dimensions are treated as independent control axes. Each axis has its own set of Conditions and Actions, allowing the agent to reason separately about horizontal geometry (e.g., lane selection, offsets) and vertical intent (e.g., climbs, descents). Manoeuvres may combine actions across both axes, but the underlying representation preserves this separation, which simplifies causal attribution, supports local plan modifications, and enables mechanisms such as monotonic axis constraints. By isolating which axis contributes to a conflict, the agent can adjust only the relevant dimension of the plan while leaving the other unchanged, reducing unintended interactions and keeping the overall plan interpretable.

\subsubsection*{Planned Actions}
A \textit{Planned Action} binds these primitives into a closed-loop unit:
\[
\mathcal{P} = \langle C_{\text{trigger}}, A, C_{\text{completion}} \rangle.
\]
The agent issues $A$ only when $C_{\text{trigger}}$ is satisfied, and monitors $C_{\text{completion}}$ to verify execution, enabling detection of deviations such as delayed pilot response or failure to achieve a commanded profile.

\subsubsection*{Manoeuvre Synthesis}
Planned Actions are composed into \textit{Manoeuvres}—structured, parameterised sequences implementing high-level strategies (e.g., lateral or vertical separation). This mirrors the hierarchical task decomposition observed in expert ATCOs \citep{seamster1993cognitive}, allowing the agent to reason at the level of tactical intent rather than individual control inputs.

For example, a \textit{Lateral Offset Manoeuvre} comprises up to three phases: \textit{Parallel} (direct both aircraft to parallel offset lanes), \textit{Maintain} (hold offset until separation threshold exceeded), and optionally \textit{Resume} (rejoin nominal routes). The Parallel phase triggers immediately on the current airspace state, while subsequent phases use geometric triggers (e.g., ``AircraftPassedLaterally'') rather than fixed timings. The manoeuvre is parameterised, in this case, by lane assignments for each aircraft (e.g., both left, both right, or opposing offsets) and the separation threshold for resumption, determined by the conflict geometry and available airspace. 

\subsubsection*{Flight Plans: Sequencing Manoeuvres Entry to Exit}

Planned Actions compose into Manoeuvres, and Manoeuvres in turn compose into \textit{Flight Plans}—the complete per-aircraft control sequence from sector entry to exit. A Flight Plan is represented as:
\[
FP = (\mathcal{P}_1 \rightarrow \mathcal{P}_2 \rightarrow \ldots \rightarrow \mathcal{P}_N)
\]
where each $\mathcal{P}_i$ is a Planned Action, and the completion condition of one action serves as the trigger for the next, forming a temporally ordered chain of logic-gated clearances.

The agent's planning process begins with the construction of a \textit{Nominal Flight Plan} for every controlled aircraft. This baseline plan produces the operationally efficient path from sector entry to the pre-coordinated exit condition. Leveraging the systemised airspace structure, the nominal plan instructs the aircraft to intercept its assigned PBN route, ascend to (or maintain) its Preferred Flight Level (PFL) to minimise fuel burn, and execute a Continuous Descent Operation (CDO) calculated to meet the exit fix constraint (a coordinated Flight Level) as late as possible. This nominal plan serves as the reference intent—the trajectory the aircraft would follow if no conflicts existed.

\subsubsection*{Airspace Plan: Sector-Wide Strategy}

At the highest level, Mallard maintains an \textit{Airspace Plan}: the complete set of all Flight Plans active in the sector simultaneously:
\[
S = \{FP_1, FP_2, \ldots, FP_M\}
\]
where $M$ is the number of aircraft currently under control or expected to enter within the next 15 minutes. The airspace plan represents the coordinated state of the entire sector: a conflict-free solution requires that all $M$ flight plans, when simulated forward together, maintain safe separation under all uncertainty scenarios.

When the forward simulation identifies a conflict between two aircraft, Mallard does not discard their Flight Plans and re-optimise from scratch. Instead, using a local repair mechanism called \textit{Plan Splicing} (detailed in Section~\ref{sec:splicing}), Mallard inserts corrective Manoeuvres into only the conflicting segments of the affected Flight Plans. Because each plan is represented as an ordered sequence of Planned Actions rather than a fixed coordinate list, the agent can inject a deconfliction Manoeuvre (e.g., a Lateral Offset) into the chain while preserving upstream and downstream logic. For example, to resolve a conflict during cruise, the agent identifies the specific Planned Actions responsible (e.g., ``Maintain PFL on centreline route'') and splices in an Offset Manoeuvre, temporarily displacing the aircraft laterally while maintaining all other aspects of the original plan.

\begin{figure}[ht]
    \centering
    \includegraphics[width=0.9\textwidth]{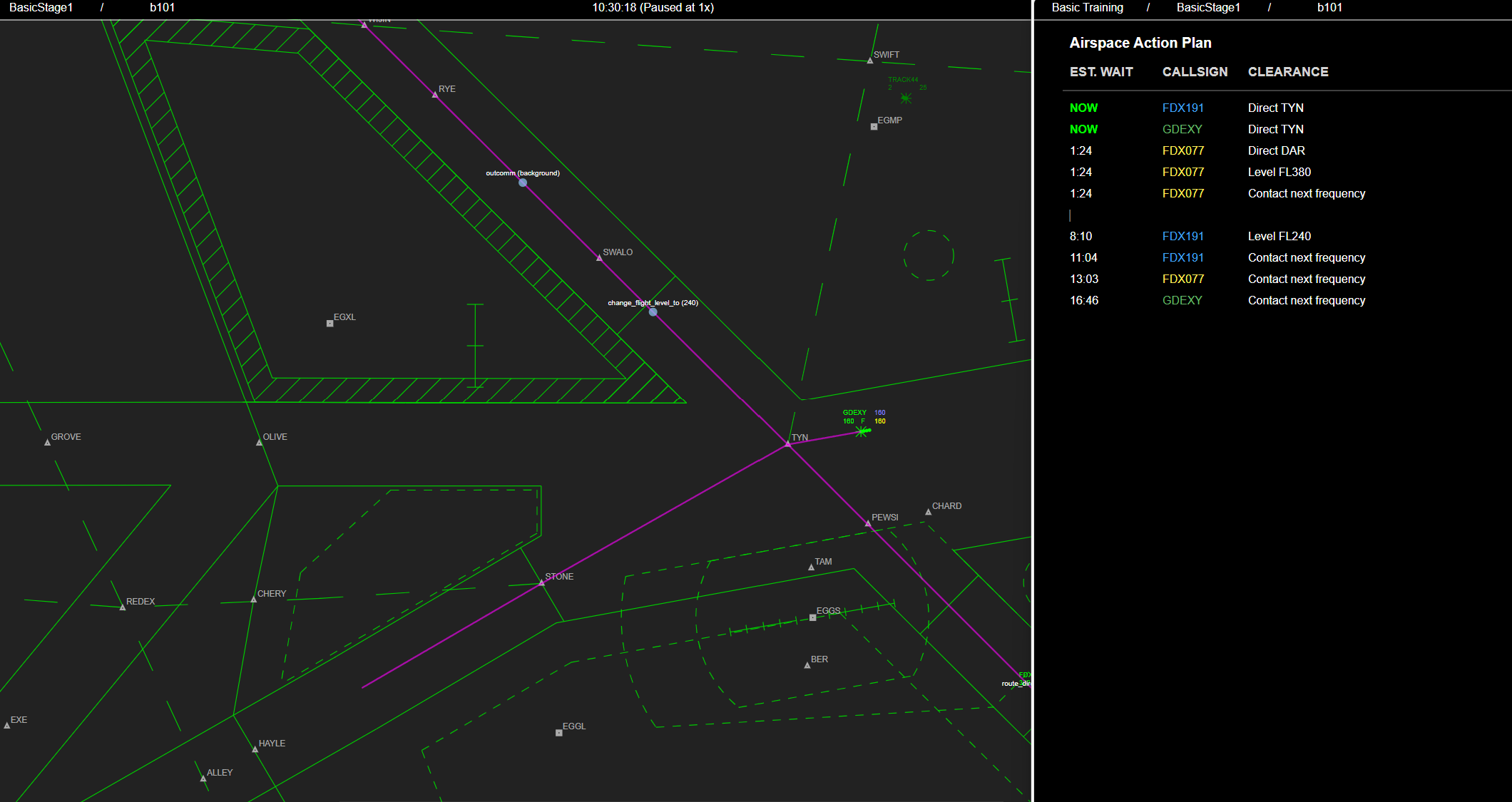}%
    \caption{Example visualisation of Agent Mallard's initial flight plans within the BluebirdDT digital twin interface, showing intended routes based on systemised airspace principles.}
    \label{fig:mallard-plan}
\end{figure}

Figure~\ref{fig:mallard-plan} shows a radar display from the BluebirdDT simulator, with multiple aircraft following systemised routes under an initial plan. These baseline rules guide each aircraft along its intended path, and if conflicts are predicted, Mallard's deconfliction logic (Section~\ref{sec:conflict_resolution}) adjusts the plans accordingly.

\subsection{Conflict Resolution}\label{sec:conflict_resolution}
Mallard continuously simulates all aircraft plans forward, flagging as a pairwise conflict any pairs of trajectories that violate separation minima or would lead to a violation if radio communication with all aircraft in the sector were lost.  We call trajectories resulting from loss of communication ``counterfactual trajectories'', as they represent alternative futures, what \textit{would} happen if communication were lost. In this instance aircraft continue to follow their route offset by whatever lane they're in.

\subsubsection{Sequential Resolution Logic}

During simulation, Mallard compiles a \textit{Technical Safety Record} (TSR) for all aircraft. Conflicts are resolved sequentially: the earliest predicted conflict is addressed first, after which the airspace is re-simulated to update downstream interactions. Re-simulation prevents the solver from correcting conflicts that no longer exist or overlooking new ones introduced by the intervention.

Conflicts are considered pairwise and are classified by both aircraft's relative vertical state, heading, and speed at the predicted Closest Point of Approach (CPA), in either nominal or counterfactual trajectories. Similarly, solutions are also pairwise and offer complementary actions to resolve.

\subsubsection{Conflict classification}

\begin{figure}[t]
\centering

\tikzset{
    vertarrow/.style={-{Latex}, thick, color=blue!70!black},
    vertaxis/.style={thin, gray, ->},
    latarrow/.style={-{Latex[length=3mm]}, thick, color=black},
    lataxis/.style={thin, gray!70, ->, >=stealth},
    spdnormal/.style={-{Latex[length=3mm]}, thick, color=black!60},
    spdfast/.style={-{Latex[length=3mm, sep=1pt] Latex[length=3mm]}, thick, color=red!80!black}
}

\begin{subfigure}{\textwidth}
    \centering
    \textbf{A. Relative Vertical Intent (Side View)}\par\medskip
    
    \begin{minipage}{0.23\textwidth}
    \centering
    \begin{tikzpicture}[scale=0.9]
      \draw[vertaxis] (-1.6,-0.5) -- (-1.6,0.5) node[above, font=\tiny]{FL};
      \draw[vertarrow] (-1.3,0) -- (1.3,0);
      \draw[vertarrow] (-1.3,-0.4) -- (1.3,-0.4);
    \end{tikzpicture}
    \subcaption{Level (LL)}
    \end{minipage}
    \hfill
    \begin{minipage}{0.23\textwidth}
    \centering
    \begin{tikzpicture}[scale=0.9]
      \draw[vertaxis] (-1.6,-0.5) -- (-1.6,0.5) node[above, font=\tiny]{FL};
      \draw[vertarrow] (-1.3,0) -- (1.3,0);
      \draw[vertarrow] (-1.3,-0.4) -- (1.3,0.4);
    \end{tikzpicture}
    \subcaption{Ascending (LA)}
    \end{minipage}
    \hfill
    \begin{minipage}{0.23\textwidth}
    \centering
    \begin{tikzpicture}[scale=0.9]
      \draw[vertaxis] (-1.6,-0.5) -- (-1.6,0.5) node[above, font=\tiny]{FL};
      \draw[vertarrow] (-1.3,0) -- (1.3,0);
      \draw[vertarrow] (-1.3,0.4) -- (1.3,-0.4);
    \end{tikzpicture}
    \subcaption{Descending (LD)}
    \end{minipage}
    \hfill
    \begin{minipage}{0.23\textwidth}
    \centering
    \begin{tikzpicture}[scale=0.9]
      \draw[vertaxis] (-1.6,-0.5) -- (-1.6,0.5) node[above, font=\tiny]{FL};
      \draw[vertarrow] (-1.3,-0.4) -- (1.3,0.4);
      \draw[vertarrow] (-1.3,0.4) -- (1.3,-0.4);
    \end{tikzpicture}
    \subcaption{Opposite (AD)}
    \end{minipage}
\end{subfigure}

\vspace{0.5cm}

\begin{subfigure}{\textwidth}
    \centering
    \textbf{B. Relative Headings (Top-Down View)}\par\medskip

    \begin{minipage}{0.3\textwidth}
    \centering
    \begin{tikzpicture}[scale=0.9]
      \draw[lataxis] (-2.0,-0.5) -- (-2.0,0.2) node[above, font=\tiny]{y};
      \draw[lataxis] (-2.0,-0.5) -- (-1.4,-0.5) node[right, font=\tiny]{x};
      \draw[latarrow] (1.6,0) -- (0.1,0);
      \draw[latarrow] (-1.6,0) -- (-0.1,0);
    \end{tikzpicture}
    \subcaption{Head-on (HO)}
    \end{minipage}
    \hfill
    \begin{minipage}{0.3\textwidth}
    \centering
    \begin{tikzpicture}[scale=0.9]
      \draw[lataxis] (-2.0,-1.2) -- (-2.0,-0.5) node[above, font=\tiny]{y};
      \draw[lataxis] (-2.0,-1.2) -- (-1.4,-1.2) node[right, font=\tiny]{x};
      \draw[latarrow] (-1.4,-0.8) -- (0,0.6);
      \draw[latarrow] (1.4,-0.8) -- (0,0.6);
    \end{tikzpicture}
    \subcaption{Crossing (CR)}
    \end{minipage}
    \hfill
    \begin{minipage}{0.3\textwidth}
    \centering
    \begin{tikzpicture}[scale=0.9]
      \draw[lataxis] (-2.0,-0.8) -- (-2.0,-0.1) node[above, font=\tiny]{y};
      \draw[lataxis] (-2.0,-0.8) -- (-1.4,-0.8) node[right, font=\tiny]{x};
      \draw[latarrow] (-1.6,0.4) -- (1.3,0.4);
      \draw[latarrow] (-1.6,-0.4) -- (1.3,-0.4);
    \end{tikzpicture}
    \subcaption{Parallel (P)}
    \end{minipage}
\end{subfigure}

\vspace{0.5cm}

\begin{subfigure}{\textwidth}
    \centering
    \textbf{C. Relative Speeds (Velocity Magnitude)}\par\medskip

    \begin{minipage}{0.3\textwidth}
    \centering
    \begin{tikzpicture}[scale=0.9]
      \draw[spdnormal] (-1.3,0.3) -- (1.3,0.3);
      \draw[spdnormal] (-1.3,-0.3) -- (1.3,-0.3);
    \end{tikzpicture}
    \subcaption{Similar ($V_1 \approx V_2$)}
    \end{minipage}
    \hfill
    \begin{minipage}{0.3\textwidth}
    \centering
    \begin{tikzpicture}[scale=0.9]
      \draw[spdfast] (-1.3,0.3) -- (1.3,0.3); 
      \draw[spdnormal] (-1.3,-0.3) -- (1.3,-0.3);
    \end{tikzpicture}
    \subcaption{AC1 Faster ($V_1 > V_2$)}
    \end{minipage}
    \hfill
    \begin{minipage}{0.3\textwidth}
    \centering
    \begin{tikzpicture}[scale=0.9]
      \draw[spdnormal] (-1.3,0.3) -- (1.3,0.3);
      \draw[spdfast] (-1.3,-0.3) -- (1.3,-0.3);
    \end{tikzpicture}
    \subcaption{AC2 Faster ($V_2 > V_1$)}
    \end{minipage}
\end{subfigure}

\caption{Comprehensive Conflict Classification Taxonomy. The agent categorises interactions across three dimensions: (A) Vertical State (Blue/Side View), (B) Lateral Geometry (Black/Plan View), and (C) Speed Differential (Red/Kinetic View). This multidimensional classification enables precise selection of deconfliction strategies.
\label{fig:conflict_taxonomy}}
\end{figure}
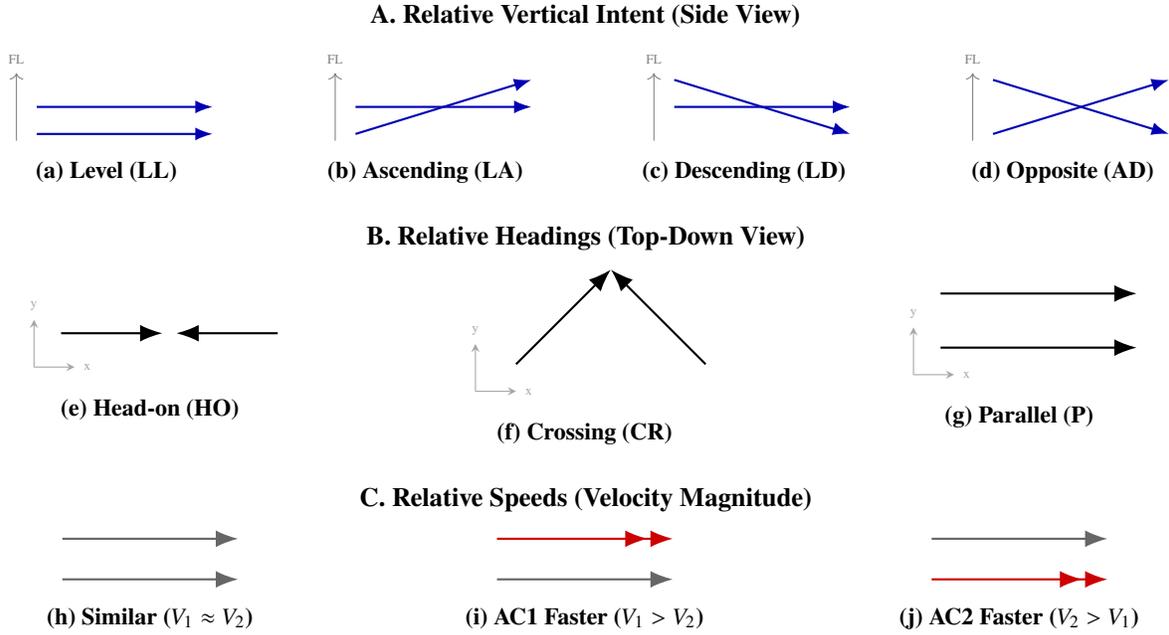

Agent Mallard classifies conflicts by integrating the following three dimensions: relative vertical state (4 combinations), relative heading (3 types), and relative speed (3 states), see Figure ~\ref{fig:conflict_taxonomy}. This creates a comprehensive taxonomy of conflict classes (e.g., \textit{Parallel / Level / AC1 Faster} denotes an overtake or catch-up). For each specific class, the agent accesses a pre-ranked list of deconfliction manoeuvres—for instance, overtake scenarios prioritise speed control or a temporary lane offset, whereas a head-on conflict prioritises vertical separation or a lateral lane change.

\subsubsection{Lane-based deconfliction}
\begin{figure}[hbtp] %
    \centering %

    \begin{subfigure}[b]{0.32\textwidth} %
        \includegraphics[width=\textwidth]{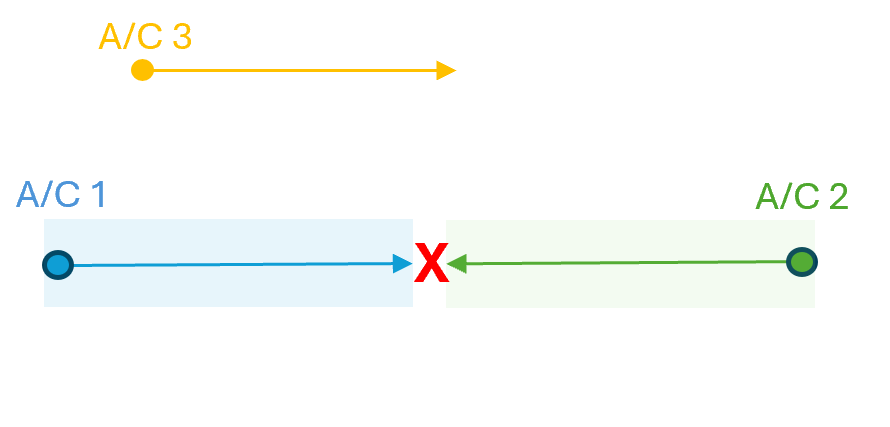} %
        \caption{} %
        \label{fig:panel_a}
    \end{subfigure}
    \hfill %
    \begin{subfigure}[b]{0.32\textwidth}
        \includegraphics[width=\textwidth]{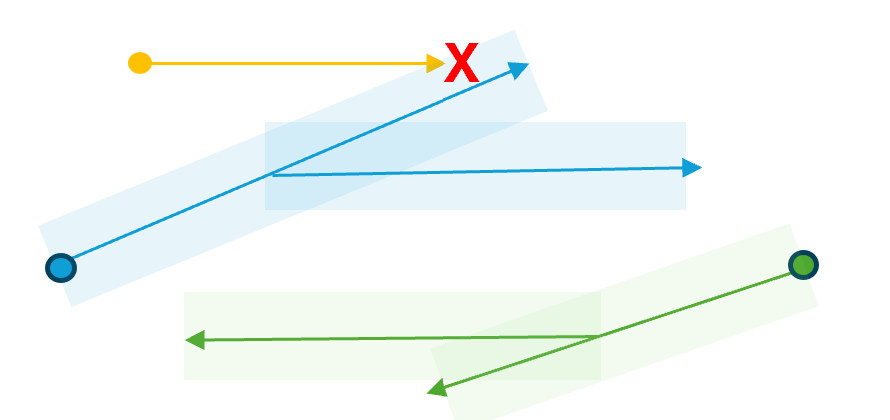} %
        \caption{} %
        \label{fig:panel_b}
    \end{subfigure}
    \hfill %
    \begin{subfigure}[b]{0.32\textwidth}
        \includegraphics[width=\textwidth]{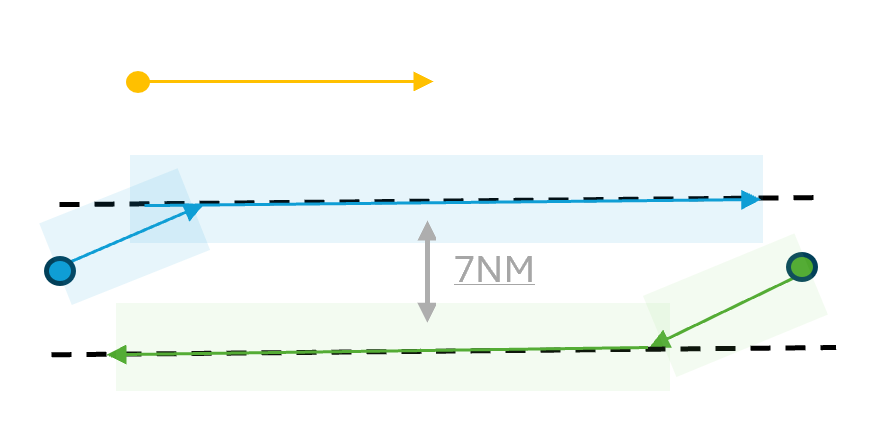} %
        \caption{} %
        \label{fig:panel_c}
    \end{subfigure}
\caption{(a) Predicted conflict (red `X') between blue and green aircraft.
(b) Tactical vectoring introduces non-standard paths and secondary conflicts (e.g., with yellow aircraft).
(c) `Dual lane change' shifts both aircraft to predefined parallel lanes, maintaining predictable separation. See main text for discussion. }
\label{fig:deconfliction_comparison}
\end{figure}

Systemised airspace, characterised by predefined PBN-guided lanes (e.g., RNAV1 which ensures that aircraft remain within 1~NM of their designated route), enables structured and predictable deconfliction. A key advantage is illustrated in Figure~\ref{fig:deconfliction_comparison}: in a head-on conflict (Figure~\ref{fig:panel_a}), a `dual lane change' (Figure~\ref{fig:panel_c}) safely transfers both aircraft to parallel lanes with known separation. With appropriate lane spacing (e.g., 7~NM between centrelines), RNAV1 guarantees aircraft will maintain at least 5~NM separation throughout.

Lane-based deconfliction provides a significant safety advantage over tactical vectoring because aircraft remain under continuous PBN guidance. Tactical vectoring (Figure~\ref{fig:panel_b}) produces open-ended headings that, in the event of communication loss, may lead aircraft onto uncontrolled paths, potentially outside sector boundaries or into secondary conflicts. In contrast, lane changes keep aircraft on predefined routes with known geometry and guaranteed endpoints. If communication is lost, the aircraft simply continues along its assigned PBN lane to the next fix, remaining within controlled airspace and preserving separation by design. This structured behaviour enables far more efficient use of lateral space and substantially simplifies the conflict-resolution problem.

\subsubsection{Airspace-Wide Conflict Resolution via Backtracking Search}\label{airspaceplan}

Although each individual deconfliction targets a specific pairwise safety issue, Mallard seeks a safe plan for all aircraft in the controlled sector.
This requirement introduces significant algorithmic complexity because  resolving one conflict may create, eliminate, or modify others elsewhere in the airspace. Consequently, the agent must account for interdependent conflict chains and coordinate resolutions across multiple aircraft pairs.

\paragraph{Foundational Mechanisms}

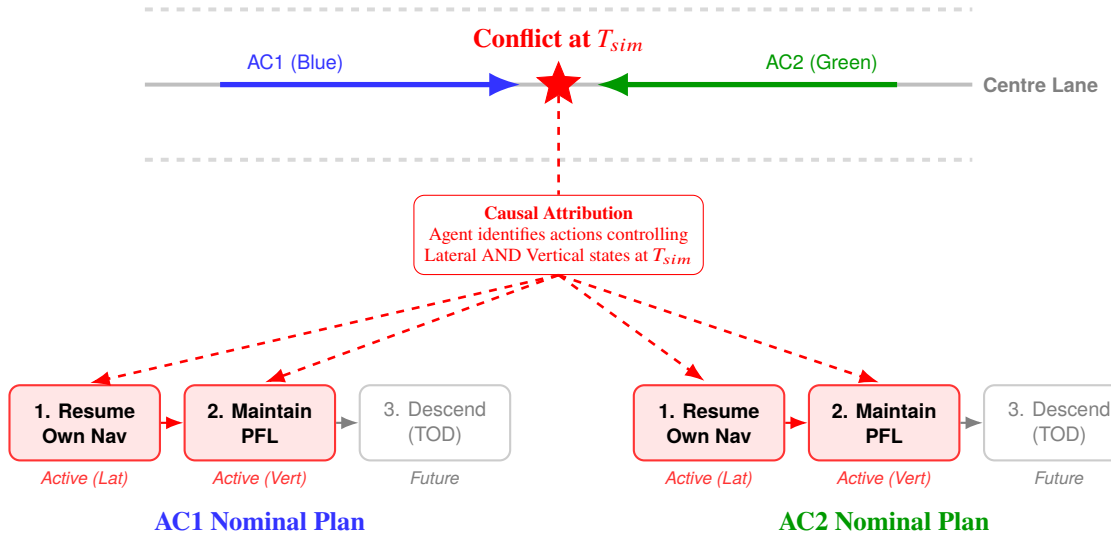
\begin{figure}[tbp]
    \centering
    \begin{tikzpicture}[
        font=\sffamily\footnotesize,
        >=Latex,
        lane/.style={gray!30, line width=1.5pt, dashed},
        traj_blue/.style={blue!80, line width=2pt, ->},
        traj_green/.style={green!60!black, line width=2pt, ->},
        active_node/.style={rectangle, draw=red!80, fill=red!10, thick, rounded corners, minimum width=2.0cm, minimum height=1.0cm, align=center},
        future_node/.style={rectangle, draw=gray!40, fill=white, thick, rounded corners, minimum width=2.0cm, minimum height=1.0cm, align=center, text=gray},
        conflict/.style={star, star points=5, star point ratio=2.25, draw=red, fill=red, inner sep=2pt, minimum size=6mm}
    ]

    \draw[lane] (-5.5, 1) -- (5.5, 1);
    \draw[gray!50, line width=1.5pt] (-5.5, 0) -- (5.5, 0) node[right, gray]{\textbf{Centre Lane}};
    \draw[lane] (-5.5, -1) -- (5.5, -1);

    \draw[traj_blue] (-4.5, 0) -- node[above, near start]{AC1 (Blue)} (-0.5, 0);
    \draw[traj_green] (4.5, 0) -- node[above, near start]{AC2 (Green)} (0.5, 0);

    \node[conflict, label={[red, font=\bfseries]above:Conflict at $T_{sim}$}] (crash) at (0,0) {};

    \node[draw=red, text=red, fill=white, rounded corners, align=center, font=\scriptsize, inner sep=4pt] (attrib) at (0, -2.0) {\textbf{Causal Attribution}\\Agent identifies actions controlling\\Lateral AND Vertical states at $T_{sim}$};

    \draw[dashed, red, line width=1pt] (crash.south) -- (attrib.north);

    \node[active_node, label={[red!80]below:\scriptsize \textit{Active (Lat)}}] (ac1_nav) at (-6.3, -4.5) {\textbf{1. Resume}\\ \textbf{Own Nav}};
    \node[active_node, right=0.3cm of ac1_nav, label={[red!80]below:\scriptsize \textit{Active (Vert)}}] (ac1_pfl) {\textbf{2. Maintain}\\ \textbf{PFL}};
    \node[future_node, right=0.3cm of ac1_pfl, label={[gray]below:\scriptsize \textit{Future}}] (ac1_tod) {3. Descend\\(TOD)};
    
    \draw[->, thick, red] (ac1_nav) -- (ac1_pfl);
    \draw[->, thick, gray] (ac1_pfl) -- (ac1_tod);

    \node[blue!80, font=\bfseries] at ($(ac1_nav.south)!0.5!(ac1_tod.south) + (0,-0.8)$) {AC1 Nominal Plan};

    \node[active_node, label={[red!80]below:\scriptsize \textit{Active (Lat)}}] (ac2_nav) at (2.0, -4.5) {\textbf{1. Resume}\\ \textbf{Own Nav}};
    \node[active_node, right=0.3cm of ac2_nav, label={[red!80]below:\scriptsize \textit{Active (Vert)}}] (ac2_pfl) {\textbf{2. Maintain}\\ \textbf{PFL}};
    \node[future_node, right=0.3cm of ac2_pfl, label={[gray]below:\scriptsize \textit{Future}}] (ac2_tod) {3. Descend\\(TOD)};
    
    \draw[->, thick, red] (ac2_nav) -- (ac2_pfl);
    \draw[->, thick, gray] (ac2_pfl) -- (ac2_tod);

    \node[green!60!black, font=\bfseries] at ($(ac2_nav.south)!0.5!(ac2_tod.south) + (0,-0.8)$) {AC2 Nominal Plan};

    \draw[->, dashed, red, line width=1pt, shorten >=2pt] (attrib.south) -- (ac1_nav.north);
    \draw[->, dashed, red, line width=1pt, shorten >=2pt] (attrib.south) -- (ac1_pfl.north);
    
    \draw[->, dashed, red, line width=1pt, shorten >=2pt] (attrib.south) -- (ac2_nav.north);
    \draw[->, dashed, red, line width=1pt, shorten >=2pt] (attrib.south) -- (ac2_pfl.north);

    \end{tikzpicture}
    \caption{\textbf{Causal Attribution and Search Space Pruning.} The Digital Twin detects a head-on conflict. The agent traces this back to the \textit{Active} actions controlling the current state. These nodes are flagged for modification (red), while Future/Pending actions (grey) are left untouched. Note, no speed actions are issued in a nominal plan.}
    \label{fig:causal_attribution}
\end{figure}

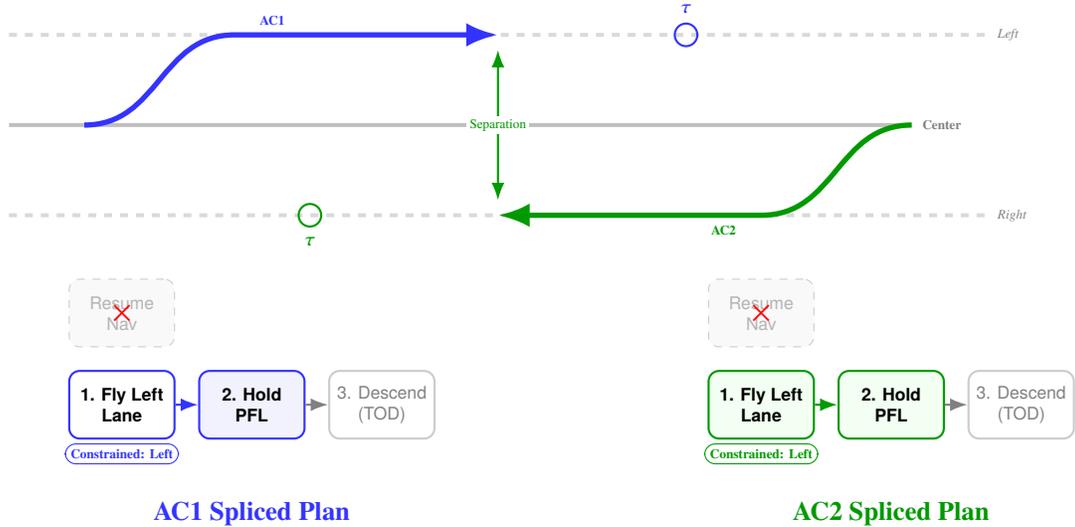
\begin{figure}[tbp]
    \centering
    \begin{tikzpicture}[
        font=\sffamily\scriptsize,
        >=Latex,
        lane/.style={gray!30, line width=1.5pt, dashed},
        traj_blue/.style={blue!80, line width=2pt, ->},
        traj_green/.style={green!60!black, line width=2pt, ->},
        active_node/.style={rectangle, draw=blue!80, fill=blue!5, thick, rounded corners, minimum width=1.4cm, minimum height=0.9cm, align=center, inner sep=2pt},
        new_node/.style={rectangle, draw=blue!80, fill=white, thick, rounded corners, minimum width=1.4cm, minimum height=0.9cm, align=center, inner sep=2pt},
        ghost_node/.style={rectangle, draw=gray!40, fill=gray!5, dashed, rounded corners, minimum width=1.4cm, minimum height=0.9cm, align=center, text=gray!60, inner sep=2pt},
        future_node/.style={rectangle, draw=gray!40, fill=white, thick, rounded corners, minimum width=1.4cm, minimum height=0.9cm, align=center, text=gray, inner sep=2pt},
        constraint_tag/.style={draw=blue!80, fill=white, text=blue!80, font=\tiny\bfseries, inner sep=2pt, rounded corners, anchor=north},
        tau_marker/.style={circle, draw, thick, inner sep=0pt, minimum size=3mm}
    ]

    \draw[lane] (-6.5, 1.2) -- (6.5, 1.2) node[right, gray, font=\tiny]{\textit{Left}};
    \draw[gray!50, line width=1.5pt] (-6.5, 0) -- (5.5, 0) node[right, gray, anchor=west, font=\tiny]{\textbf{Center}};
    \draw[lane] (-6.5, -1.2) -- (6.5, -1.2) node[right, gray, font=\tiny]{\textit{Right}};

    \draw[traj_blue] (-5.5, 0) .. controls (-4.5, 0) and (-4.5, 1.2) .. (-3.5, 1.2) -- (0.0, 1.2);
    \node[blue!80, font=\bfseries\tiny] at (-3.0, 1.4) {AC1};
    
    \node[tau_marker, draw=blue!80, label={[blue!80]above:$\tau$}] at (2.5, 1.2) {};

    \draw[traj_green] (5.5, 0) .. controls (4.5, 0) and (4.5, -1.2) .. (3.5, -1.2) -- (0.0, -1.2);
    \node[green!60!black, font=\bfseries\tiny] at (3.0, -1.4) {AC2};

    \node[tau_marker, draw=green!60!black, label={[green!60!black]below:$\tau$}] at (-2.5, -1.2) {};

    \draw[<->, thick, green!60!black] (0, 1.0) -- (0, -1.0) node[midway, fill=white, inner sep=1pt, font=\tiny]{Separation};

    \node[ghost_node] (ac1_ghost) at (-5.0, -2.5) {Resume\\Nav};
    \node[red, font=\bfseries\large] at (ac1_ghost.center) {$\times$};

    \node[new_node, below=0.3cm of ac1_ghost] (ac1_offset) {\textbf{1. Fly Left}\\ \textbf{Lane}};
    
    \node[active_node, right=0.3cm of ac1_offset] (ac1_pfl) {\textbf{2. Hold}\\ \textbf{PFL}};
    
    \node[future_node, right=0.3cm of ac1_pfl] (ac1_tod) {3. Descend\\(TOD)};

    \draw[->, thick, blue!80] (ac1_offset) -- (ac1_pfl);
    \draw[->, thick, gray] (ac1_pfl) -- (ac1_tod);
    
    \node[constraint_tag, below=0.05cm of ac1_offset] {Constrained: Left};

    \node[blue!80, font=\bfseries, below=0.7cm of ac1_pfl] {AC1 Spliced Plan};

    \node[ghost_node] (ac2_ghost) at (3.5, -2.5) {Resume\\Nav};
    \node[red, font=\bfseries\large] at (ac2_ghost.center) {$\times$};

    \node[new_node, draw=green!60!black, fill=green!5, below=0.3cm of ac2_ghost] (ac2_offset) {\textbf{1. Fly Left}\\ \textbf{Lane}};

    \node[active_node, draw=green!60!black, fill=green!5, right=0.3cm of ac2_offset] (ac2_pfl) {\textbf{2. Hold}\\ \textbf{PFL}};
    
    \node[future_node, right=0.3cm of ac2_pfl] (ac2_tod) {3. Descend\\(TOD)};

    \draw[->, thick, green!60!black] (ac2_offset) -- (ac2_pfl);
    \draw[->, thick, gray] (ac2_pfl) -- (ac2_tod);

    \node[constraint_tag, text=green!60!black, draw=green!60!black, below=0.05cm of ac2_offset] {Constrained: Left};

    \node[green!60!black, font=\bfseries, below=0.7cm of ac2_pfl] {AC2 Spliced Plan};

    \end{tikzpicture}
    \caption{The conflicting ``Resume Nav'' action (crossed out) is bypassed. A new instruction (Fly Left Lane) is spliced into the sequence, constraining the lateral axis. The subsequent vertical logic (Hold PFL) and future logic (Descend) are preserved. $\tau$ indicates the future point at which the lateral constraint is released giving that aircraft the freedom to accept alternative deconfliction strategies on that axis.}
    \label{fig:plan_splicing}
\end{figure}

Before describing the complete resolution algorithm, we introduce three key mechanisms that enable effective conflict resolution:

\begin{enumerate}[label=\Roman*., ref=\thesubsubsection.\Roman*]
    \item \textbf{Causal Attribution (Search Space Pruning):}
    \label{sec:causal}

When a conflict is detected, Mallard performs causal attribution by tracing the trajectory backward to identify the \textit{Planned Action} which is immediately responsible for the unsafe state (Figure~\ref{fig:causal_attribution}). Isolating this causal element allows the agent to modify only the relevant segment of the plan while preserving the rest. Once a safe revision is found, the original later actions remain valid, reducing unnecessary changes and narrowing the search space.

    \item \textbf{Plan Splicing:}
    \label{sec:splicing}

Once a valid strategy is selected, Mallard applies it via \textit{Plan Splicing} (Figure~\ref{fig:plan_splicing}). The sequenced planned action structure allows the agent to replace only the conflicting segment with the new manoeuvre, itself a sequence of planned actions. In the example in the figure a complementary replacement of the lateral planned action in both aircraft plans solves the conflict.

    \item \textbf{Monotonic Axis Constraints:}
    \label{sec:monotonic}

To reduce oscillatory behaviour and encourage convergence to stable plans, Mallard applies \textit{Monotonic Axis Constraints}. Once an intervention modifies a particular axis, for example issuing an \textit{Offset Right}, that axis becomes constrained. Any subsequent intervention may only reinforce that decision (e.g., increase the offset) and may not reverse it. This applies independently to each axis of control. Each constraint has an associated completion condition indicating when the underlying safety issue has passed. In Figure~\ref{fig:plan_splicing}, the $\tau$ markers show the point at which the aircraft have passed one another and the lateral constraint is lifted. By preventing contradictory manoeuvres, monotonicity eliminates cycling between incompatible fixes and steers the search toward a stable solution.

\end{enumerate}

\paragraph{Algorithm Structure: Sequential Resolution with Backtracking}\label{sec:backtracking}

With these foundational mechanisms established (Figure~\ref{fig:causal_attribution} 
and Figure~\ref{fig:plan_splicing}), we now describe how Mallard coordinates them into a complete resolution strategy. Mallard proposes a depth-limited backtracking search algorithm that explores ranked strategies for each conflict and coordinates sequential conflict resolution. Let $S_0$ denote the initial Airspace Plan composed of nominal Flight Plans for all aircraft in the sector, and let $X_0$ denote the current airspace state as observed from surveillance. The algorithm is formalised in Algorithm~\ref{alg:backtracking}.

\begin{algorithm}[h!]
\caption{Airspace-Wide Conflict Resolution}
\label{alg:backtracking}
\begin{algorithmic}[1]
\State \textbf{Input:} Current plan set $S$, Current airspace state $X$, max depth $d_{max} = 3$
\State \textbf{Output:} Complete safe plan set $S'$ or fallback
\State
\Function{ResolveAirspace}{$S$, $X$, $depth$}
    \State $conflicts \gets \textsc{SimulateAndDetect}(S, X)$ \Comment{Forward-simulate all aircraft}
    
    \If{$conflicts = \emptyset$}
        \State \Return $S$ \Comment{Success: safe plan set found}
    \EndIf
    
    \If{$depth \geq d_{max}$}
        \State \Return \textsc{null} \Comment{Depth limit exceeded}
    \EndIf
    
    \State $c \gets \textsc{EarliestConflict}(conflicts)$ \Comment{Temporal priority}
    \State $\mathcal{S}_{strat} \gets \textsc{GetStrategies}(c)$
    \State $\mathcal{S}_{strat} \gets \textsc{FilterByAxisConstraints}(\mathcal{S}_{strat})$
    
    \ForAll{$s \in \mathcal{S}_{strat}$} \Comment{Try strategies in priority order}
        \State $S' \gets \textsc{ApplyStrategy}(S, s, c)$ \Comment{Plan splicing on causal segment (Section~\ref{sec:splicing})}
        \State $solution \gets \textsc{ResolveAirspace}(S', X, depth + 1)$ \Comment{Recursive call}
        
        \If{$solution \neq \textsc{null}$}
            \State \Return $solution$ \Comment{First complete safe plan wins}
        \EndIf
        
        \State \textsc{UndoStrategy}(S, s, c) \Comment{Backtrack: restore previous plans}
    \EndFor
    
    \State \Return \textsc{null} \Comment{All strategies exhausted}
\EndFunction
\State
\State \textbf{Main:}
\State $solution \gets \textsc{ResolveAirspace}(S_0, X_0, 0)$
\If{$solution = \textsc{null}$}
    \State \Return \textsc{FallbackStrategy}() \Comment{Conservative vertical separation}
\EndIf
\State \Return $solution$
\end{algorithmic}
\end{algorithm}

\paragraph{Operational Flow}

The algorithm operates through coordinated cycles of strategy application and airspace-wide validation. The digital twin simulates all aircraft plans across the 1-hour planning window and identifies predicted separation losses, which are sorted by time to form a temporal priority queue. The earliest conflict is addressed first to maintain causal consistency, ensuring the agent does not fix conflicts that later interventions would eliminate. The search proceeds with a bounded recursion depth, where depth $d$ counts the number of sequential conflict resolutions applied since the initial airspace state. The algorithm begins at depth $d=0$ with the baseline Nominal Plans. For the selected conflict at the current depth, candidate resolution strategies from the library (Table~\ref{tab:strategy_library}) are considered in priority order. Causal attribution (Section~\ref{sec:causal}) identifies the responsible plan segment, and monotonic axis constraints (Section~\ref{sec:monotonic}) prune incompatible options before each strategy is evaluated.

Each candidate strategy is applied via plan splicing (Section~\ref{sec:splicing}), modifying only the causal segment. The entire airspace is then re-simulated to detect cascading effects, such as new conflicts or changes in geometry. If conflicts remain, the algorithm recurses to depth $d+1$, resolving the next predicted conflict under the updated state.  

If no strategy at the current depth resolves all downstream conflicts, or if the depth limit $d_{\max}=3$ is reached, the algorithm backtracks: the most recent plan modification is undone, and the next strategy at depth $d$ is attempted. Exhausting all strategies at a given depth triggers further backtracking to depth $d-1$ until a complete safe plan is found or all options fail. It should be noted that the maximum depth was chosen during testing and in principle the depth can be as large as required.

\paragraph{Strategy Ordering with Greedy Acceptance}

The algorithm combines two  design principles:

\textit{Strategy Ordering} At each conflict, strategies from the  library (Table~\ref{tab:strategy_library}) are considered in priority order. The search depth limit discourages excessive cascading interventions. In practice, this priority order means that simpler solutions (requiring fewer interventions) tend to be found earlier, aligning with the operational preference for low-complexity plans.\looseness=-1

\textit{Greedy First-Solution Acceptance:} The search terminates as soon as any complete safe airway plan is found. The algorithm does not exhaustively enumerate all possible solutions to select the globally optimal one. This greedy acceptance provides significant computational savings: once a valid plan is discovered, no further backtracking or strategy exploration occurs.

For example, if Strategy 1 (lateral separation, as example) resolves a head-on conflict at depth d=1 without creating new conflicts, the algorithm accepts this immediately even if Strategy 3 might theoretically produce a slightly shorter path.

Because strategies are evaluated in priority order and the search is depth-limited, the search tends to find shallow solutions that use high priority strategies, aligning with the operational preference for low-complexity plans. Among the solutions explored before termination, the first complete safe plan will typically be one of the shallowest in terms of intervention depth, although it is not guaranteed to be a globally minimal-depth or globally optimal solution.

\paragraph{Computational Properties}

\textit{Best Case:} If each conflict resolves cleanly with its highest-ranked strategy (no secondary conflicts introduced), the algorithm exhibits $O(k )$ complexity, where $k$ is the number of initial conflicts. In practice, well-designed strategy priorities increase the probability of this best-case scenario: the highest priority strategy often succeeds without creating follow-on complications.

\textit{Worst Case:} In adversarial scenarios where every strategy creates new conflicts requiring backtracking, the complexity approaches $O(b^{d_{max}})$ where $d_{max}$ is the maximum search depth and $b$ is the branching factor (number of strategies per conflict class). However, this exponential worst case is mitigated by several factors: (1) the depth limit ($d_{max} = 3$) bounds the search, (2) axis constraints prune many branches, (3) most real-world conflicts admit simple resolutions, and (4) greedy acceptance prevents exhaustive search.

\textit{Termination Guarantee:} The algorithm is guaranteed to terminate. The depth bound ensures that recursion cannot proceed indefinitely, and the finite strategy libraries ensure that each conflict node has bounded branching factor. If no safe airway plan exists within the search space, the algorithm exhausts all possibilities and invokes the fallback mechanism.

\paragraph{Fallback Mechanism}

If the search exhausts all strategy combinations up to depth $d_{max}$ without finding a complete safe airway plan, Mallard invokes a conservative fallback: an immediate vertical separation manoeuvre (identifying unoccupied flight levels for both aircraft and sending them there) designed to provide a simple, deterministic safety buffer. The system simultaneously alerts human controllers, flagging the scenario as an edge case requiring expert intervention. The intent of this fail-safe is to prioritise separation in highly constrained scenarios beyond the agent's resolution capability, while explicitly preserving human supervisory authority. In our tests, this fallback is considered a failure and is simply an attempt to approximate a reasonable failure recovery mode.

\begin{table}[H]
\centering
\caption{\textbf{Deconfliction Strategy Library.} After identifying the causal actions, the agent selects a resolution by evaluating known strategies in priority order. Each strategy is tested through forward simulation before commitment.}
\label{tab:strategy_library}
\begin{tabular}{clp{8.5cm}}
\toprule
\textbf{Priority} & \textbf{Strategy } & \textbf{Description} \\
\midrule
1 & \textsc{Lateral separation (Same side)} & Used for head-on directions at the same level by assigning lateral offsets on the same side of centreline. \\
2 & \textsc{Level Change (Descent)} & One aircraft descends below the conflict flight level. \\
3 & \textsc{Level Change (Climb)} & One aircraft climbs above the conflict flight level; used when the lower level is unavailable. \\
4 & \textsc{Exchange Levels} & Aircraft separate laterally, pass each other vertically, and then rejoin their original lateral route. \\
5 & \textsc{Match speed and trail} & Set both aircraft on speed control and trail one behind the other on the same track.\\
6 & \textsc{Additional Strategies} & Additional strategies or parameterised variants. \\
\bottomrule
\end{tabular}
\end{table}

The backtracking search algorithm provides a mechanism for assessing safety through forward simulation.
However, safety alone is insufficient for operational acceptance. Controllers must be able to understand, verify, and appropriately trust the agent's reasoning. The following section describes how Mallard's architectural design enables this transparency.

\subsection{Interpretability and Human–Machine Teaming}\label{sec:hmi}
A  limitation of many proposed AI  ATC agents is their opacity~\citep{EASA2020AIRoadmap, IFATCA2022AIChallengesATM}. Trust in such systems is shaped not only by successful outcomes
but also by the transparency of the underlying decision process~\citep{bluebird_agent_transparency}. Mallard addresses this through architectural design: the hierarchical plan representation (Section \ref{sec:hierarchy}) preserves decision provenance, causal attribution (Section \ref{sec:causal}) identifies responsible plan segments, and the complete resolution algorithm (Algorithm \ref{alg:backtracking}) maintains strategy selection history. To operationalise this transparency, we have developed visualisation tools designed to give insight into the algorithm's operation and reasons for the choice of resolution strategies. %
\subsubsection{Visualisation for Training and Validation}
\begin{figure}
    \centering
    \includegraphics[width=0.9 \linewidth]{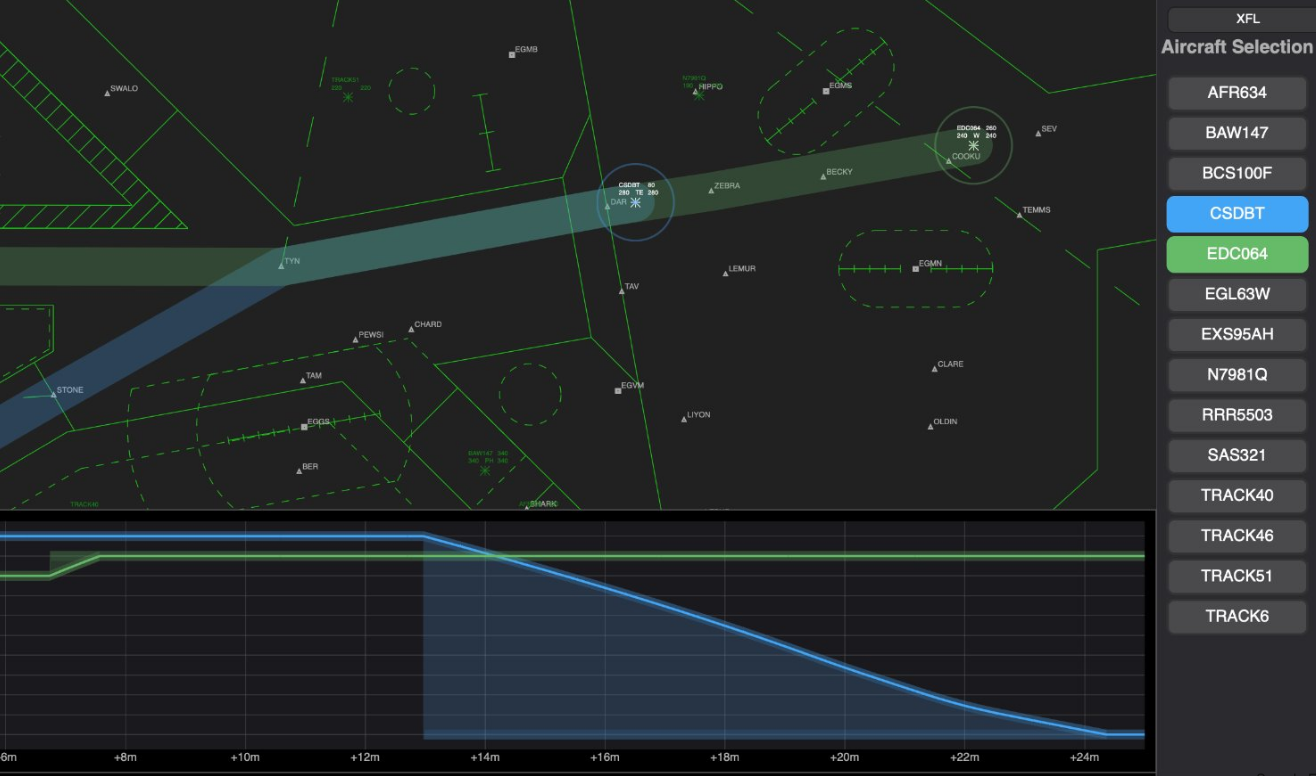}
    \caption{A visualisation of a conflict from the human machine interface (HMI). Lateral positions (with safety buffer) appear on the radar screen, vertical intent is shown on the lower pane. Two conflicting aircraft are automatically selected, and their plans show an unsafe procedure in around 13 minutes. At this time, the blue aircraft will descend through the green aircraft and their lateral positions will be unsafe. Controllers can scrub through the timeline to see the trajectories and confirm the safety issue.}
    \label{fig:conflict_viz}
\end{figure}

To support ATCO familiarisation and agent validation, we propose a detailed visualisation interface that exposes Mallard's internal reasoning at multiple levels of abstraction:
\begin{itemize}
\item \textbf{Airspace Plan View}: Sector-wide display of all aircraft trajectories (Figure \ref{fig:mallard-plan}), showing nominal plans and any spliced deconfliction segments, allowing controllers to inspect the complete intended state of the airspace.
\item \textbf{Per-Aircraft Plan Inspector}: Detailed view of the Planned Action sequence for selected aircraft, showing the condition-action-completion chain that governs each phase of the flight.

\item \textbf{Conflict Resolution History}: A temporal, scrubbable visualisation linking each detected conflict to its classification, the strategy selected from the ranked library (Table \ref{tab:strategy_library}), and the resulting plan modifications. This trace is designed to allow controllers to replay the agent's decision-making process step-by-step (see Figure \ref{fig:conflict_viz}).
\end{itemize}
These detailed views are intended for offline analysis: understanding the agent's reasoning patterns during training scenarios, identifying edge cases where strategy selection may be suboptimal, and building controller trust through transparency. The scrubbable conflict history is designed to support post-scenario review, allowing instructors and controllers to examine how cascading conflicts were resolved and whether the agent's strategy choices align with expert judgement.

While this detailed transparency is essential for validation and trust-building, presenting this level of detail during routine operations would impose unnecessary cognitive burden.
\subsubsection{Minimal Display for Operational Use}
For real-time human-machine teaming, we propose an intentionally minimal operational interface. Controllers would see:
\begin{itemize}
\item Proposed trajectories for all aircraft with anticipated action points indicating where trigger conditions will be satisfied and clearances issued
\item Clearances ready for controller review and issuance
\item Conflict alerts with suggested resolutions
\end{itemize}
This operational view is designed to focus on \emph{intent} rather than reasoning: controllers can verify that planned trajectories maintain separation and align with sector coordination requirements, without needing to inspect the underlying strategy selection logic during routine operations. The detailed visualisation tools would remain available if controllers wish to query the agent's rationale for a particular decision, but would not be presented by default.
Prototype implementations of both visualisation modes have been developed within the BluebirdDT environment, but empirical evaluation of their effectiveness with controllers remains future work (Section \ref{sec:future_hmi}).
This two-mode design recognises that interpretability serves different purposes in different contexts: transparent validation for building trust and competence assessment, versus minimal intent display for maintaining situational awareness without cognitive overload during real-time operations.

With the complete methodology established, from trajectory prediction through conflict resolution to decision transparency, we now describe how Mallard's performance will be systematically assessed.

\section{Validation Methodology}\label{sec:validation}

Agent performance will be evaluated using the Machine Basic Training (MBT) ~\citep{bluebird_validation}, a human-centric evaluation methodology developed under Project Bluebird and based on the rigorous training and assessment standards used for human Air Traffic Controller (ATCO) trainees at the NATS College. The MBT assesses agents across four operational dimensions aligned with ATCO training standards:

\begin{itemize}
    \item \textbf{Safety}: Maintenance of separation minima under realistic 
    execution uncertainty
    \item \textbf{Planning}: Strategy quality, operational impact minimisation, 
    downstream conflict prevention
    \item \textbf{Coordination}: Aircraft state at sector boundaries, handover 
    clarity
    \item \textbf{Interpretability}: Alignment with standard phraseology, logical 
    coherence
\end{itemize}

Evaluation scenarios are drawn from ATCO training material within the BluebirdDT 
digital twin (Section~\ref{sec:digital_twin_description}), covering varying traffic densities, conflict geometries, and cascading multi-aircraft interactions. Expert ATCO instructors review agent-generated solutions for operational acceptability, human-likeness, and strategic appropriateness.

\section{Current Status and Preliminary Validation}
\label{sec:current_status}

Agent Mallard is currently in active development, with core architectural components implemented and beginning to undergo initial validation in the BluebirdDT environment. Preliminary assessment has focused on establishing the operational credibility of the approach:

\subsection{Logic Verification with Expert ATCOs}
The hierarchical manoeuvre structures and conflict resolution strategies have been reviewed through structured walkthroughs with active UK controllers and ATCO instructors. These sessions involved:
\begin{itemize}
    \item Presenting the conflict classification system and ranked strategy libraries
    \item Walking through example scenarios on the simulator to verify logical coherence
    \item Soliciting feedback on manoeuvre parametrisation and completion conditions
    \item Assessing the interpretability of the hierarchical plan representation
\end{itemize}

Controller feedback has been encouraging: in trial scenarios, experts indicated that Mallard's decision logic broadly mirrors familiar operational reasoning patterns and that the proposed manoeuvres are operationally sound for the presented conflict geometries. Several refinements to strategy ranking and parametrisation have emerged from these consultations.

\subsection{Partial Implementation and Testing}
Key components of the Mallard architecture have been implemented:
\begin{itemize}
    \item Hierarchical plan representation (Conditions, Actions, Planned Actions, Manoeuvres)
    \item Digital twin integration for trajectory simulation
    \item Conflict detection and classification logic
    \item Core deconfliction strategy library for common conflict classes
\end{itemize}

Early tests on a small subset of MBT scenarios have provided an initial indication that the simulation-in-the-loop approach can detect conflicts, generate candidate manoeuvres, and resolve them without introducing secondary violations under the digital twin model. A systematic MBT-based evaluation campaign remains future work.

\subsection{Next Steps: Full MBT Evaluation}
Complete validation through the MBT framework is ongoing, with the following objectives:
\begin{enumerate}
    \item \textbf{Scenario Coverage:} Systematic testing across the full MBT syllabus
    \item \textbf{Quantitative Metrics:} Collection of safety, efficiency, and computational performance data
    \item \textbf{Expert Assessment:} Formal review of agent solutions by ATCO instructors
    \item \textbf{Comparative Analysis:} Benchmarking against human controller performance on matched scenarios
    \item \textbf{Failure Mode Analysis:} Identification of boundary conditions and edge cases
\end{enumerate}

\section{Design Assumptions and Known Limitations}
\label{sec:assumptions_limitations}

Agent Mallard's current design operates under key assumptions that define its scope and recognise areas for future development:

\paragraph{Systemised Airspace Context}
Mallard is designed for tactical control within well-defined systemised airspace where aircraft follow PBN-guided routes. The agent assumes:
\begin{itemize}
    \item Aircraft comply with issued clearances and maintain RNAV1 performance
    \item Lane structures remain geometrically valid and appropriately spaced
    \item Nominal operational conditions (no severe weather, system failures, or airspace structure invalidation)
\end{itemize}

In scenarios where these assumptions break down (e.g., large-scale weather disruptions requiring free routing), the current Mallard design would not be applied as-is; instead, control would revert to human ATCOs or to alternative tools better suited to free route operations.

\paragraph{Pairwise Conflict Resolution}
The core deconfliction logic is optimised for pairwise encounters. While the architecture handles some cascading conflicts through iterative resolution, highly dense multi-aircraft cluster scenarios may exceed current capabilities. This represents a known boundary for the present implementation.

\paragraph{Expert-Informed Strategy Space}
As a rules-based system informed by expert knowledge, Mallard's strategic repertoire reflects encoded ATCO practices. It does not generate novel manoeuvre types beyond its programmed library. This design choice prioritises interpretability and operational alignment over unconstrained solution exploration.

\paragraph{Simulation Fidelity Dependency}
Performance relies on the accuracy and timeliness of the digital twin's aircraft state data, trajectory models, and environmental conditions. Degraded data quality or modelling errors would propagate to agent decisions.

\paragraph{Search Termination and Fallback}
If no safe solution is found from among the strategies, the system flags the conflict as requiring human intervention and issues a 'safe flight level' manoeuvre intended to move both aircraft promptly to a safer flight level, explicitly abandoning coordinated exits or flight-economy considerations.

\section{Future Work}\label{sec:future_work}

Agent Mallard is in active development. The following areas represent priorities 
for advancing the system towards operational evaluation and deployment:

\subsection{Comprehensive Validation}

The most immediate priority is systematic evaluation through the complete MBT scenario syllabus~\citep{bluebird_validation}. The MBT provides a structured curriculum of operational scenarios, each representing approximately thirty minutes of simulated traffic through a training sector. Scenarios are designed with specific assessment criteria to systematically build agent capabilities, progressing from fundamental conflict resolution through increasingly complex tactical situations. Current testing covers a small subset of foundational scenarios with isolated conflict geometries. Complete validation requires:

\begin{itemize}
    \item \textbf{Progressive Complexity:} Systematic progression through the full MBT curriculum, addressing scenarios with multiple concurrent conflicts, cascading interactions, and constrained solution spaces
    
    \item \textbf{Operational Realism:} Testing under realistic sector traffic flows rather than constructed pairwise encounters, including normal entry/exit patterns, varying aircraft performance, and mixed flight profiles (climb, cruise, descent)
    
    \item \textbf{Assessment Criteria Coverage:} Demonstration of competence across all MBT assessment dimensions—Safety (separation assurance under uncertainty), Planning (strategy quality and efficiency), Coordination (exit state management), and Interpretability (operational clarity of solutions)
    
    \item \textbf{Quantitative Performance Metrics:} Collection of separation assurance rates, trajectory efficiency (fuel burn, time deviation, path stretch), computational performance, and strategy selection distributions across the complete scenario set

    \item \textbf{Formal Expert Assessment:} Structured review of agent-generated solutions by ATCO instructors, comparing Mallard's approach to human controller solutions on matched scenarios
\end{itemize}

This systematic evaluation will identify performance boundaries, failure modes, 
and scenarios requiring strategy library extension or algorithmic refinement 
before operational deployment can be considered.

\subsection{Strategy Library Extension and Expertise Elicitation}
The current strategy library (Table 2) covers fundamental conflict classes identified through initial expert consultations. However, operational expertise is distributed across controllers with varying experience, sector specialisations, and regional practices. Systematic extension of the library through structured ATCO engagement could serve both as a practical refinement mechanism and as a window into the diversity of tactical reasoning in air traffic control.

\subsubsection*{Proposed Elicitation Approach:} As Mallard is evaluated on MBT scenarios, controllers could observe Mallard's proposed solutions and provide structured feedback: accepting the approach as operationally sound, or offering alternative strategies they would prefer in the same situation. Because Mallard's hierarchical plan representation is interpretable—explicitly showing conflict classification, strategy selection, and parametrised manoeuvre sequences—controllers can meaningfully critique not just outcomes but decision rationale. This would enable strategy discovery (identifying valid approaches not currently in the library and documenting the conditions under which controllers prefer alternatives), parametrisation refinement (observing how controllers adjust manoeuvre parameters in specific contexts), strategy ranking calibration (documenting where expert preferences diverge from current rankings and why), and recognition of context-dependent adaptation rules (patterns in how strategies should be modified for specific operational scenarios—exit coordination, flow management, neighbouring sector constraints, weather, or temporary restrictions).

\subsubsection*{Potential Research Insights:} This iterative refinement process could yield broader insights into the structure of tactical expertise. By systematically documenting where and why controllers' preferences differ from the codified library, such an approach might help characterise: the degree of consensus versus variation in tactical decision-making across the controller population; whether strategy selection is primarily driven by conflict geometry (suggesting universal principles) or operational context (suggesting sector-specific expertise); and the relative importance of different constraints (safety margins, efficiency, coordination complexity, workload) in shaping strategy preferences. Mallard's interpretability is essential to this process: because the agent exposes its reasoning at the level of conflict classification, strategy selection, and manoeuvre decomposition, controllers can provide feedback at the appropriate level of abstraction, adjusting high-level strategy choices rather than low-level control parameters. This positions strategy library development not merely as system improvement but potentially as a methodology for studying expert tactical reasoning in safety-critical domains.

\subsection{Algorithmic Enhancements}

Exploitation of modern multicore architectures to evaluate multiple candidate strategies simultaneously would reduce wall-clock resolution time. This is entirely compatible with the depth-limited backtracking search already defined.

\subsection{Human–Machine Collaboration Protocols}\label{sec:future_hmi}

While the visualisation architecture described in Section \ref{sec:hmi} supports transparency, systematic empirical evaluation remains future work. Key open questions include: optimal information density for operational use and controller intervention protocols when overriding agent suggestions. These should be investigated through structured user studies combining objective performance metrics with subjective trust and workload assessments.

The ultimate objective is not to replace human controllers but to provide reliable, transparent decision support for routine tactical deconfliction, enabling controllers to allocate attention to strategic coordination, off-nominal situations, and high-level flow management where human judgement remains essential.

\section*{Acknowledgments}
This work was supported by the EPSRC (EP/V056522/1) and NATS.

\clearpage
\bibliography{ref}

\end{document}